\documentclass{article}

\usepackage{graphicx}
\usepackage{multirow}
\usepackage{amsmath,amssymb,amsfonts}
\usepackage{amsthm}
\usepackage{mathrsfs}
\usepackage[title]{appendix}
\usepackage{xcolor}
\usepackage{textcomp}
\usepackage{booktabs}
\usepackage{algorithm}
\usepackage{algorithmicx}
\usepackage{algpseudocode}
\usepackage{listings}
\usepackage{makecell}
\usepackage{array}
\usepackage{url}
\newcolumntype{C}[1]{>{\centering\arraybackslash}m{#1}}

\begin{document}

\title{Multi-View MRI Approach for Classification of MGMT Methylation in Glioblastoma Patients}

\author{
Rawan Alyahya\thanks{National Center for Artificial Intelligence (NCAI), SDAIA, Riyadh, Saudi Arabia} \and
Asrar Alruwayqi\footnotemark[1] \and
Atheer Alqarni\footnotemark[1] \and
Asma Alkhaldi\footnotemark[1] \and
Metab Alkubeyyer\footnotemark[1] \and
Xin Gao\thanks{King Abdullah University of Science and Technology (KAUST), CEMSE Division, Thuwal, Saudi Arabia} \and
Mona Alshahrani\thanks{Corresponding author: mona.alshahrani@kaust.edu.sa \\ 
Aramco Research Center, R\&D Center Department, Dhahran, Saudi Arabia}
}
\date{}

\maketitle


\begin{abstract}
The presence of MGMT promoter methylation significantly affects how well chemotherapy works for patients with Glioblastoma Multiforme (GBM). Currently, confirmation of MGMT promoter methylation relies on invasive brain tumor tissue biopsies. In this study, we explore radiogenomics techniques, a promising approach in precision medicine, to identify genetic markers from medical images. Using MRI scans and deep learning models, we propose a new multi-view approach that considers spatial relationships between MRI views to detect MGMT methylation status. Importantly, our method extracts information from all three views without using a complicated 3D deep learning model, avoiding issues associated with high parameter count, slow convergence, and substantial memory demands. We also introduce a new technique for tumor slice extraction and show its superiority over existing methods based on multiple evaluation metrics. By comparing our approach to state-of-the-art models, we demonstrate the efficacy of our method. Furthermore, we share a reproducible pipeline of published models, encouraging transparency and the development of robust diagnostic tools. Our study highlights the potential of non-invasive methods for identifying MGMT promoter methylation and contributes to advancing precision medicine in GBM treatment.
\end{abstract}


\section{Introduction}

Gliomas are the most common brain malignancies and occur in different grades and levels of severity \cite{louis20072007}. The World Health Organization (WHO) estimates that 60--70\% of malignant gliomas are grade IV glioblastoma multiforme (GBM), the most aggressive tumors of the central nervous system (CNS) \cite{louis20072007}. Standard treatments include surgical resection followed by radiation therapy and Temozolomide chemotherapy \cite{stupp2009effects}. However, despite aggressive treatment, the average overall survival remains low at 14.6 months and can drop to 5.6 months for patients aged 80 and above \cite{johnson2012glioblastoma}.

Studies have shown that methylation of the O\(^6\)-methylguanine--DNA methyltransferase (MGMT) repair enzyme is a strong indicator of chemotherapy response in GBM patients. Patients with methylated MGMT tend to respond better to Temozolomide and radiotherapy and have improved prognosis \cite{zhang2013prognostic,wen2008malignant}. According to \cite{hegi2005mgmt}, patients with methylated MGMT are 45\% more likely to survive, with a median survival of 18.2 months compared to 12.2 months for unmethylated cases. Therefore, non-invasive identification of MGMT methylation status is crucial for optimizing treatment decisions.

Magnetic resonance imaging (MRI) is widely used for brain tumor assessment due to its excellent soft tissue contrast and ability to capture three-dimensional structural information \cite{ellingson2014pros,weizman2012prediction}. The Brain Tumor Segmentation (BraTS) Challenge, organized in collaboration with ASNR and MICCAI, provides a benchmark platform for developing machine learning models for tumor segmentation and classification. For the first time, BraTS 2021 includes labels for MGMT methylation status \cite{baid2021rsna}.

In this work, we propose a fully automated framework for classifying MGMT methylation status. Our main contributions are as follows:

\begin{itemize}
\item We propose a multi-view model for MGMT methylation classification along with a new tumor slice extraction technique that significantly improves AUROC performance compared to state-of-the-art models and single-view approaches.
\item We reduce 3D MRI volume classification to a 2D classification problem, effectively lowering computational cost while achieving superior performance.
\item We release a reproducible pipeline for existing approaches, facilitating transparent and accessible benchmark analysis for this challenging task.
\end{itemize}

The source code and reproducible pipeline are available at:
\url{https://github.com/rawanalyahya/Multi-View-MRI-Approach-for-Classification-of-MGMT-Methylation-in-Glioblastoma-Patients}

\section{Related Work}\label{sec2}

Existing research related to MGMT methylation status classification has predominantly followed two approaches: radiomics-based approaches and MRI-based approaches. The first approach is based on studies showing a correlation between MGMT methylation status and MRI-extracted features \cite{eoli2007methylation,levner2009predicting}. These methods extract radiomics features from the region of interest in a brain tumor, which can be identified through a brain segmentation model \cite{gillies2016radiomics}. The extracted features are then passed to a classical machine learning model to predict the methylation status. On the other hand, MRI-based approaches use raw MRI image data and follow a deep learning approach, such as Convolutional Neural Networks (CNNs) and Recurrent Neural Networks (RNNs).

Radiomics-based feature extraction aims to compute a large number of quantitative descriptors from medical images using data characterization algorithms \cite{lambin2017radiomics}. Visually assessed features \cite{drabycz2010analysis}, texture features from 2D MRI slices \cite{korfiatis2016mri}, and VASARI features \cite{kanas2017learning} have all been examined for MGMT methylation prediction. Sveinn et al.\ \cite{palsson2021prediction} proposed combining radiomics and latent shape features to predict MGMT methylation status. They obtained latent shape features from a variational autoencoder (VAE). The authors used a total of 23 features, 16 of which were radiomics features and 7 were latent shape features. After training their model on the RSNA-ASNR-MICCAI BraTS 2021 dataset, they achieved a validation AUC score of 0.598.

Authors in \cite{li2018multiregional} evaluated a large number of radiomics features from multiple tumor subregions in multi-parametric MRI images by automatically extracting 1,705 multi-regional radiomics features. Their goal was to develop a multiregional and multiparametric radiomics model that could predict MGMT promoter methylation status in GBM before therapy. According to researchers in \cite{sasaki2019radiomics}, radiomics can also be used to produce a prognosis score that differentiates between high- and low-risk GBM, and this score was found to be predictive independent of MGMT promoter methylation status.

The MRI-based approach utilizes deep learning techniques to extract image features from sequences of MRI frames. One of the early studies in this direction was proposed by Korfiatis et al.\ \cite{korfiatis2017residual}, which employed transfer learning to predict MGMT status without localizing the tumor. They compared multiple residual deep neural network (ResNet) architectures and concluded that a 50-layer ResNet performed the best. Their study conducted classification on a slice-by-slice basis, where every slice of each patient was labeled as methylated or unmethylated and passed through a 2D ResNet model.

Han and Kamdar \cite{han2018mri} combined CNNs with RNNs to maintain information transfer across MRI slices. They presented a bidirectional CRNN model using the axial plane to predict MGMT methylation status. Chang et al.\ \cite{chang2018deep} introduced a 2D–3D hybrid CNN for predicting MGMT promoter status using MRI scans from the TCGA and TCIA datasets. They demonstrated that individual genetic mutations in both low- and high-grade gliomas can be classified.

Chen et al.\ \cite{chen2020automatic} designed an end-to-end pipeline that incorporates both glioma tumor segmentation and MGMT methylation status prediction. They achieved this by cascading a 4-layer CNN classification model with a tumor segmentation model for GBM patients. Similar to \cite{korfiatis2017residual}, they labeled and trained their data on a slice-by-slice basis, utilizing all axial slices of each patient.

A summary of existing work on MGMT methylation status is presented in the supplementary materials.

Recently, several studies have explored applying 2D models to accelerate 3D segmentation in domain-specific challenges. These are referred to as 2.5D models in the literature \cite{bermejo2018emphysema,ottesen20232,avesta2023comparing}. Such methods fuse multiple 2D segmentation outputs to generate a 3D segmented volume. They have shown improved performance in various applications, such as higher brain metastasis detection rates in MRI \cite{ottesen20232}, and improved liver and tumor segmentation in CT when compared to both 2D and 3D models \cite{li2018h}. Additionally, Li Xiaomeng et al.\ \cite{zhou2019d} demonstrated lower computational cost and improved efficiency using 2.5D models for chronic stroke lesion segmentation.

All of these MRI-based approaches rely on the entire 3D MRI volume, either through 3D models or by passing each slice individually to a 2D model. This requires high computational resources and long training times. In addition, most previous works on MGMT methylation prediction do not provide their code and are difficult to reproduce. Challenges include the lack of implementation details, the use of private datasets, or dependence on pixel-level tumor labels not provided in BraTS 2021.

We mitigate these challenges by using only three slices from each patient's 3D MRI volume. The concept of using a single slice from multiple views (a 2.5D representation) has gained traction for accelerating 3D data processing. The work in \cite{8363629} introduced a 2.5D multi-view CNN approach for emphysema classification in CT scans. Their model outperformed deeper architectures in sensitivity and specificity. Similarly, \cite{avesta2023comparing} studied 3D, 2.5D, and 2D auto-segmentation methods and found that while 3D models achieved higher Dice scores, 2.5D models had faster training and inference times. In medical imaging applications such as COVID-19 CT segmentation, \cite{zhou2020rapid} decomposed the 3D task into three 2D tasks, significantly reducing complexity while maintaining accuracy. These studies highlight the versatility and efficiency of 2.5D techniques.

In this work, we utilize a multi-view approach to predict MGMT methylation status from MRI volumes. We also facilitate future comparisons by re-implementing key state-of-the-art MGMT classification models and releasing them as executable notebooks with clear instructions.

\section{Methods}\label{sec3}

\subsection{Data Collection}\label{subsubsec1}
Data was collected from the RSNA-ASNR-MICCAI BraTS 2021 dataset, which includes glioma multiparametric MRI (mpMRI) scans taken from different scanners and vendors under various scanning protocols. This results in heterogeneous image quality, reflecting different clinical practices across facilities worldwide. The dataset contains 585 cases in total; however, case numbers 113 and 123 were excluded because they were not successfully pre-processed by the pipeline described below. As a result, the dataset size was reduced to 583 cases, split into 420 for training (220 methylated and 200 unmethylated), 47 for validation (25 methylated and 22 unmethylated), and 116 for testing (62 methylated and 54 unmethylated). Each case includes four MRI modalities: fluid attenuated inversion recovery (FLAIR), native T1-weighted (T1w), T2-weighted (T2w), and T1w contrast-enhanced (T1wCE). All MRI scans were skull-stripped, anonymized, and stored in DICOM format.

\begin{figure}
  \centering
  \includegraphics[width=0.9\textwidth]{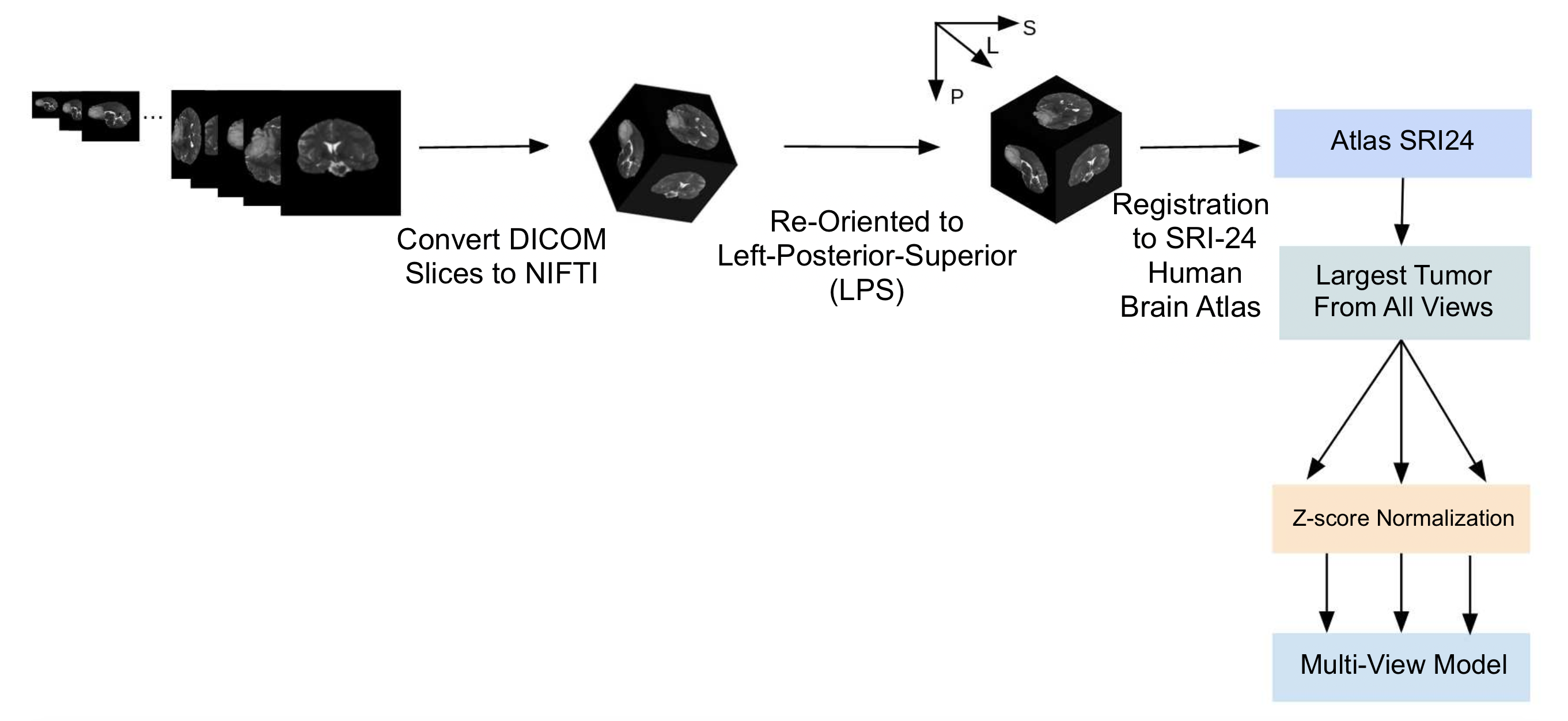}
  \caption{The proposed pipeline takes a heterogeneous collection of a patient's MRI scans with different resolutions in the DICOM format. The images then undergo pre-processing steps to produce a uniform MRI volume, which is used as the input to our multi-view model.}
  \label{fig:pipeline}
\end{figure}

\subsection{Data Pre-processing}\label{preprocessing}
The BraTS 2021 dataset exhibits heterogeneity in both anatomical orientation and slice count across cases. Therefore, a three-stage pre-processing pipeline was applied. First, DICOM images were converted to the NIFTI format for each modality (Figure~\ref{fig:pipeline}). Second, volumes were re-oriented to the left-posterior-superior (LPS) anatomical convention. Third, registration to the SRI-24 Human Brain Atlas \cite{rohlfing2010sri24} was performed to ensure a uniform anatomical coordinate system (Figure~\ref{fig:pipeline}). Bias correction was applied temporarily to facilitate registration, but the final registered image was the original (non–bias-corrected) image to maintain fidelity. The SRI-24 atlas has a $1\,\mathrm{mm}^3$ isotropic resolution with dimensions of $240 \times 240 \times 155$. After registration, all volumes matched this resolution and dimensionality, removing slice-count variability. These steps were implemented using the CaPTk toolkit \cite{pati2019cancer,davatzikos2018cancer}.

Pixel intensities of each extracted 2D image were normalized using z-score normalization:

\begin{equation}
p_{\mathrm{new}} = \frac{p - \mu}{\sigma},
\end{equation}

where $p$ is the original pixel value, $\mu$ is the mean, and $\sigma$ is the standard deviation.

To improve generalization and augment the dataset, we applied random horizontal and vertical flips, random rotation between $-10^\circ$ and $+10^\circ$, and random sharpness adjustments.

\subsection{Tumor Segmentation}
\begin{figure}
\centering
\includegraphics[width=0.65\textwidth]{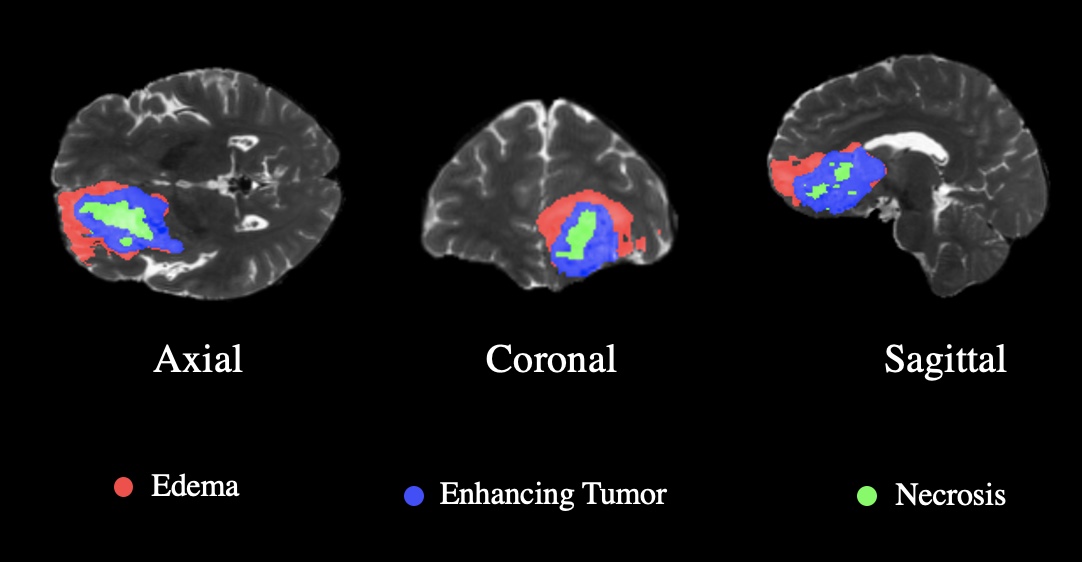}
\caption{Glioma subregions segmented by nnU-Net, shown in three anatomical views of the same brain.}
\label{fig:subreigons}
\end{figure}

Tumor segmentation was performed using a pre-trained nnU-Net model \cite{isensee2021nnu}. nnU-Net automatically configures the entire segmentation pipeline for a given dataset and adapts well to diverse biomedical segmentation tasks. The publicly available weights trained on BraTS 2020 (highly similar to BraTS 2021 \cite{baid2021rsna}) were used. All four MRI modalities (FLAIR, T1w, T2w, T1wCE) were input to the model to segment tumor subregions: edema, necrosis, and enhancing tumor (Figure~\ref{fig:subreigons}).

\subsection{Multi-View MRI Model}
\label{sec:model}

\begin{figure}[thpb]
\centering
\includegraphics[width=0.9\textwidth]{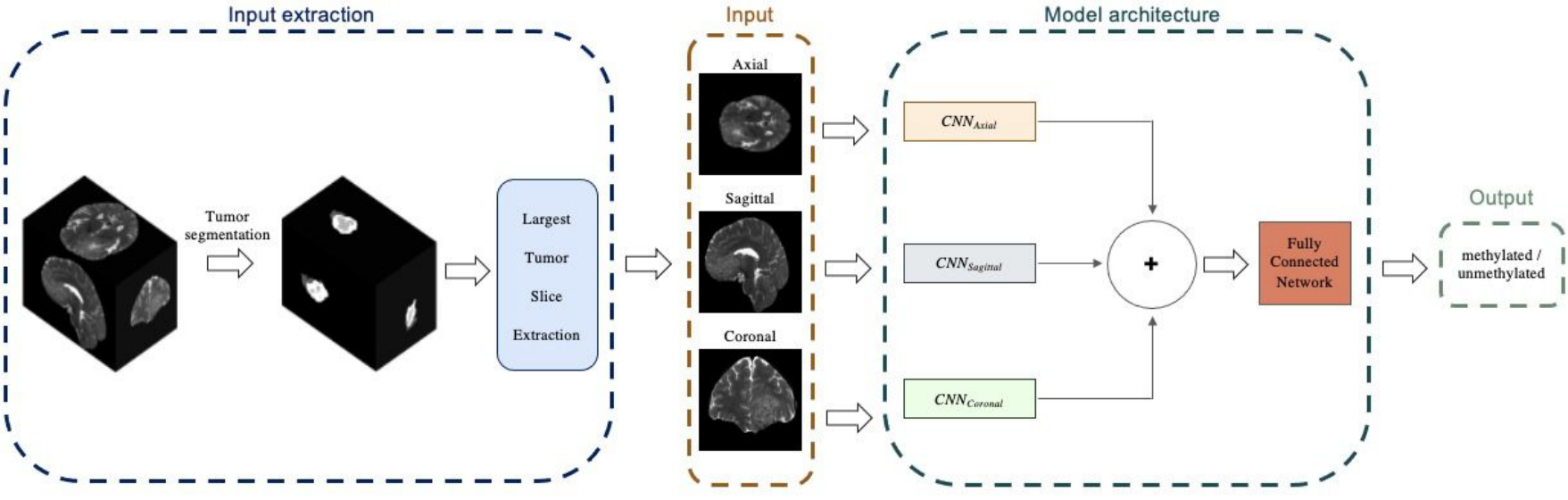}
\caption{Overview of the multi-view model. Tumor segmentation identifies slices in each anatomical plane; the slice with the maximum Feret diameter is extracted from axial, sagittal, and coronal views. Each slice is processed through a MONAI Densenet-121 branch, and extracted features are concatenated and passed to a fully connected classifier for MGMT methylation prediction.}
\label{fig:model}
\end{figure}

We propose a multi-view model that uses one slice from each MRI view (axial, sagittal, coronal). As illustrated in Figure~\ref{fig:model}, the pipeline consists of four stages. First, the segmented tumor region is used to identify the slice containing the largest tumor in each view. Second, each selected slice is passed through a CNN to extract feature maps \cite{the_monai_consortium_2020_4323059}. Third, the three feature vectors are concatenated. Finally, a fully connected network predicts MGMT methylation status.

\subsubsection{Largest Tumor Slice Extraction}
\begin{figure}
\centering
\includegraphics[width=0.8\textwidth]{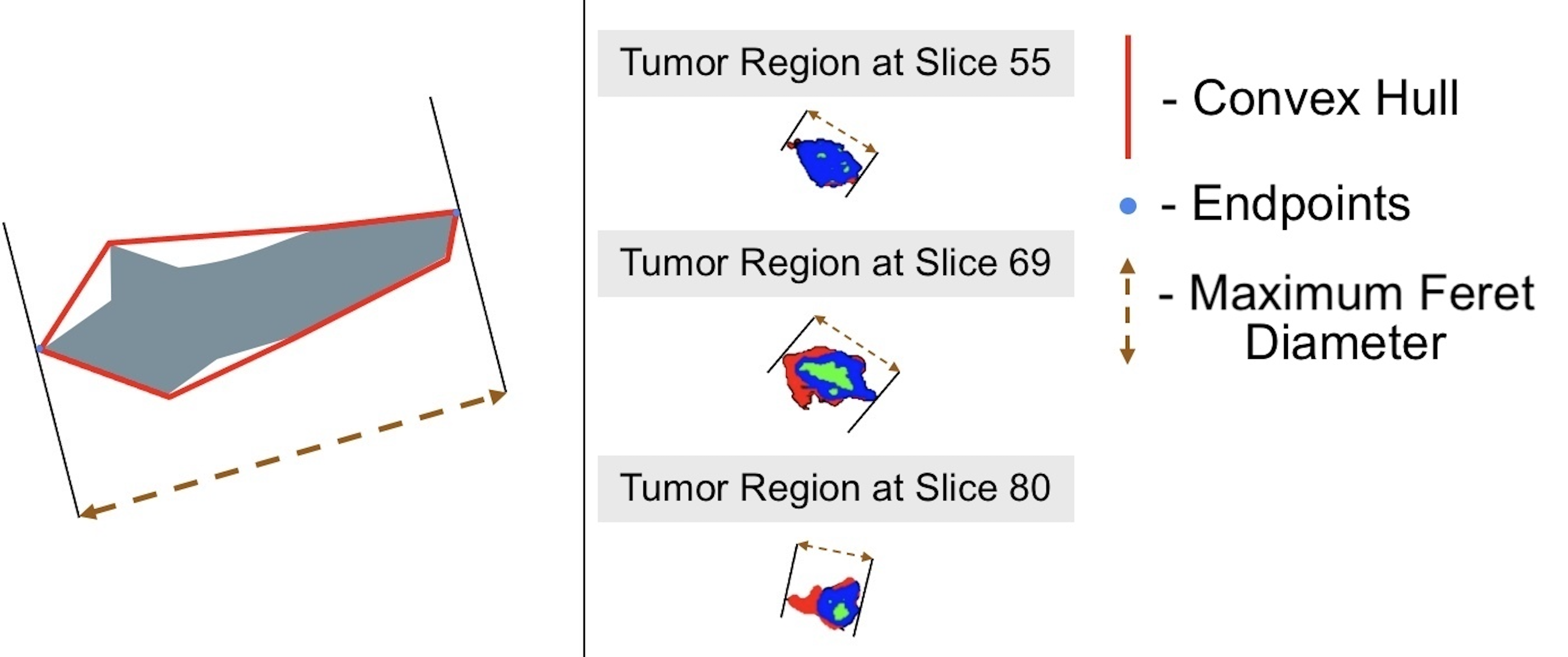}
\caption{Maximum Feret diameter computation. Left: illustration of Feret diameter as the maximum distance between two points of the convex hull. Right: example slices from one patient showing that slice 69 has the largest tumor extent.}
\label{fig:feret}
\end{figure}

Each MRI volume contains 240 sagittal slices, 240 coronal slices, and 155 axial slices. To reduce redundancy and computational cost, we select only one slice per view. The slice chosen is the one with the largest visible tumor. Using the segmented tumor mask, we compute the maximum Feret diameter for each slice: the longest distance between any two points on the convex hull of the tumor (Figure~\ref{fig:feret}). The slice with the maximum diameter is selected for each view.

We compared this method with two alternative slice-selection strategies: (1) selecting the slice with the largest tumor area (pixel count), and (2) selecting the slice with the maximum Martin diameter (the length of the area bisector). Results in Table~\ref{tab:tab1} show that the Feret-based method performed best, whereas the alternatives tended to bias predictions toward the methylated class.

\begin{table}[h]
\caption{Comparison between slice extraction methods.}
\label{tab:tab1}
\begin{tabular}{@{}lllll@{}}
\toprule
Slice extraction method & AUC & $F_1$ Score & Recall & 95\% CI \\
\midrule
Largest tumor Feret diameter & \textbf{0.662} & 0.633 & 0.613 & \textbf{0.563--0.761} \\
Largest tumor Martin diameter & 0.450 & 0.606 & 0.694 & 0.343--0.558 \\
Largest tumor area & 0.419 & \textbf{0.641} & \textbf{0.790} & 0.310--0.527 \\
\bottomrule
\end{tabular}
\end{table}

\subsubsection{Methylation Status Classification}
To predict MGMT methylation status, we use three pre-trained CNN branches, one for each view: $\mathrm{CNN}_{\mathrm{axial}}$, $\mathrm{CNN}_{\mathrm{sagittal}}$, and $\mathrm{CNN}_{\mathrm{coronal}}$, following approaches from Wu et al.\ \cite{wu2020deep} and Zhou et al.\ \cite{zhou2020rapid,zhou2022interpretable}. Each branch consists of a MONAI Densenet-121 model pre-trained on MedMNIST \cite{the_monai_consortium_2020_4323059}, chosen due to domain similarity and strong performance on medical grayscale images.

Each CNN outputs a feature vector, denoted $\vec{v}_{\mathrm{axial}}$, $\vec{v}_{\mathrm{sagittal}}$, and $\vec{v}_{\mathrm{coronal}}$. These vectors are concatenated:

\begin{equation}
\vec{v}_{\mathrm{multi\text{-}view}} = [\,\vec{v}_{\mathrm{axial}},\ \vec{v}_{\mathrm{sagittal}},\ \vec{v}_{\mathrm{coronal}}\,].
\end{equation}

The combined vector is passed through a fully connected classifier containing three layers to produce the final binary prediction. This multi-view design captures complementary spatial information across views and mitigates the limitations of small training size by leveraging pre-trained initialization.

\section{Experiments and Results}

\subsection{Implementation Details}
Our study exclusively utilized T2-weighted images (T2wi) as the input. This choice was driven by the ability of T2wi to effectively highlight glioblastoma and surrounding edema. T2-weighted images are widely used in neuro-oncology for their high sensitivity in visualizing tumor extent, which is essential for identifying MGMT methylation status. All experiments were implemented in Python using the PyTorch library. Hyperparameter tuning was conducted by experimenting with different learning rates ($10^{-7}$, $10^{-8}$, $10^{-6}$, $10^{-5}$, $10^{-4}$), batch sizes (8, 16, 32, 64), optimizers (Adam, SGD, Adagrad), and numbers of neurons in the fully connected layers. The optimal configuration based on validation performance was a learning rate of $10^{-3}$, batch size of 16, and the Adagrad optimizer. Early stopping was applied if validation accuracy did not improve for 100 epochs, with a maximum of 400 epochs.

\subsection{Multi-View Model Results}

To evaluate the performance of our multi-view model, we compared the results with single-view models using only one of the axial, coronal, or sagittal slices. We also compared our method with a 3D approach using a 3D ResNet101 pretrained on MedMNIST. Our model achieved the best performance across all metrics, including:

\begin{itemize}

\item \textbf{Area under the ROC curve (AUC)}:
\begin{equation}
\mathrm{AUC} = \int_{0}^{1} \mathrm{TPR}(\mathrm{FPR}) \, d(\mathrm{FPR})
\end{equation}

\item \textbf{Precision}:
\begin{equation}
\mathrm{Precision} = \frac{TP}{TP + FP}
\end{equation}

\item \textbf{Recall}:
\begin{equation}
\mathrm{Recall} = \frac{TP}{TP + FN}
\end{equation}

\item \textbf{Specificity}:
\begin{equation}
\mathrm{Specificity} = \frac{TN}{TN + FP}
\end{equation}

\item \textbf{$F_{1}$ score}:
\begin{equation}
F_{1} = \frac{2 \cdot \mathrm{Precision} \cdot \mathrm{Recall}}{\mathrm{Precision} + \mathrm{Recall}}
\end{equation}

\item \textbf{95\% confidence interval (CI)}
\end{itemize}

Our multi-view model achieved an AUC of 0.662, compared to 0.556 for axial, 0.553 for coronal, 0.568 for sagittal, and 0.551 for the 3D volume. This improvement is statistically significant (permutation test, $p = 0.060$). Additional results appear in Table~\ref{tab:tab2}.

\begin{table*}
\centering
\caption{Comparison of our Multi-View approach with single-view approaches using axial, coronal, and sagittal slices.}
\label{tab:tab2}
\begin{tabular}{@{}lllllll@{}}
\toprule
Input Type & AUC & $F_{1}$ Score & Recall & Precision & Specificity & 95\% CI \\ \midrule
Sagittal       & 0.568 & 0.617 & 0.661 & 0.577 & 0.444 & 0.461--0.675 \\
Coronal        & 0.553 & 0.571 & 0.548 & 0.596 & 0.574 & 0.445--0.662 \\
Axial          & 0.556 & 0.671 & \textbf{0.871} & 0.545 & 0.167 & 0.450--0.662 \\
3D Volume      & 0.551 & 0.667 & 0.839 & 0.553 & 0.222 & 0.444--0.658 \\
Multi-View     & \textbf{0.662} & \textbf{0.633} & 0.613 & \textbf{0.655} & \textbf{0.630} & \textbf{0.563--0.761} \\
\bottomrule
\end{tabular}
\end{table*}

\subsection{Comparisons with State-of-the-Art Models}

We compared our approach with several published MGMT methylation classification methods, applying the same pre-processing pipeline to ensure consistency. Table~\ref{tab:tab3} summarizes these comparisons.

\textbf{Korfiatis et al.} \cite{korfiatis2017residual} proposed a slice-wise training approach using a ResNet50. Each slice was assigned the patient's label. Our re-implementation resulted in 13{,}525 normal slices, 13{,}210 methylated slices, and 11{,}633 unmethylated slices in the training set. At test time, majority voting was used. Although \cite{korfiatis2017residual} reported around 90\% accuracy on a private dataset, our implementation achieved an AUC of 0.512 with recall = 1.0 and specificity = 0.0, meaning all samples were predicted as methylated.

\textbf{P\'alsson et al.} \cite{palsson2021prediction} proposed extracting radiomics features from tumor subregions (edema, enhancing tumor, necrosis) followed by a random forest classifier. Following their approach, we extracted features per subregion and trained 117 models via cross-validation, averaging predictions during testing. This method achieved an AUC of 0.426.

We also compared against the \textbf{Kaggle-winning 3D ResNet-10} model, which uses the ``Central Image Trick'' to unify slice counts. Using our train/val/test split, it achieved an AUC of 0.562, approximately 10\% lower than our multi-view model.

\begin{table*}
\centering
\caption{Performance comparison between the proposed Multi-View model and existing approaches for MGMT classification.}
\label{tab:tab3}
\begin{tabular}{|l|l|l|l|l|l|l|}
\hline
Approach & Input Data & AUROC & $F_{1}$ Score & Recall & Precision & Specificity \\ \hline
Slice-wise labeling \cite{korfiatis2017residual} & T2w MRI & 0.512 & \textbf{0.698} & \textbf{1.000} & 0.536 & 0.000 \\ \hline
Radiomics features \cite{li2018multiregional} & Radiomics vectors & 0.573 & 0.581 & 0.548 & 0.618 & 0.618 \\ \hline
Kaggle 3D ResNet-10 model & T2w MRI & 0.562 & 0.519 & 0.452 & 0.609 & \textbf{0.673} \\ \hline
Multi-View (proposed) & T2w MRI & \textbf{0.662} & 0.633 & 0.613 & \textbf{0.655} & 0.630 \\ \hline
\end{tabular}
\end{table*}

\section{Limitations and Future Work}

This work is limited by its reliance on a single dataset, the RSNA-ASNR-MICCAI BraTS 2021 dataset, which contains 585 cases. After applying our pre-processing steps and splitting the data into training, validation, and testing sets, only 420 cases remained for training. A larger and more diverse dataset would likely improve generalization and reduce the risk of overfitting to the characteristics of this specific cohort.

Another limitation is that our study focuses solely on the T2-weighted (T2w) MRI modality. While T2w images are highly informative for visualizing glioblastoma and surrounding edema, incorporating additional MRI modalities or exploring multi-modal fusion strategies may provide complementary information and potentially enhance MGMT methylation prediction performance.

\section{Conclusion}

In this work, we proposed a multi-view model that outperforms single-view approaches for MGMT methylation status classification. The model leverages information from the axial, coronal, and sagittal views of a 3D MRI volume by selecting, for each view, the slice containing the largest tumor region as determined by the maximum Feret diameter. This reduces the 3D classification problem to an efficient 2D framework while still incorporating rich spatial information from multiple orientations.

We also developed a unified pre-processing pipeline to standardize MRI resolution and orientation across patients, enabling consistent and reproducible experimentation. Our method achieved a significant improvement over existing baseline models on the RSNA-ASNR-MICCAI BraTS 2021 dataset. In addition, we provide a reproducible and publicly accessible pipeline, including standardized pre-processing steps and evaluation metrics, to support future research and facilitate fair comparisons across methods. This work aims to accelerate progress toward more accurate and clinically meaningful models for non-invasive prediction of MGMT methylation status.


\bibliographystyle{plain}
\bibliography{bibliography}

\end{document}